\documentclass{article}
\PassOptionsToPackage{numbers, compress}{natbib}
\usepackage[preprint]{neurips_2019}
\usepackage{hyperref}

\usepackage{natbib}
\usepackage[utf8]{inputenc}
\usepackage[T1]{fontenc}   
\usepackage{hyperref}      
\usepackage{url}            
\usepackage{booktabs}       
\usepackage{amsfonts}     
\usepackage{amsmath}      
\usepackage{nicefrac}    
\usepackage{microtype}     
\usepackage{graphicx}
\usepackage{bm}
\usepackage{amssymb}
\usepackage[flushleft]{threeparttable}
\usepackage{subfigure}
\usepackage{makecell}
\usepackage{float}

\newcommand*{\QED}{\hfill\ensuremath{\square}}%

\title{Estimation with Uncertainty via Conditional Generative Adversarial Networks}

\author{%
  Minhyeok Lee \\
  School of Electrical Engineering\\
  Korea University\\
  Seoul, South Korea \\
  \texttt{suam6409@korea.ac.kr} \\
  \And
  Junhee Seok\thanks{To whom correspondence should be addressed.} \\
  School of Electrical Engineering\\
  Korea University\\
  Seoul, South Korea \\
  \texttt{jseok14@korea.ac.kr} \\
}

\begin{document}

\maketitle

\begin{abstract}
Conventional predictive Artificial Neural Networks (ANNs) commonly employ deterministic weight matrices; therefore, their prediction is a point estimate. Such a deterministic nature in ANNs causes the limitations of using ANNs for medical diagnosis, law problems, and portfolio management, in which discovering not only the prediction but also the uncertainty of the prediction is essentially required. To address such a problem, we propose a predictive probabilistic neural network model, which corresponds to a different manner of using the generator in conditional Generative Adversarial Network (cGAN) that has been routinely used for conditional sample generation. By reversing the input and output of ordinary cGAN, the model can be successfully used as a predictive model; besides, the model is robust against noises since adversarial training is employed. In addition, to measure the uncertainty of predictions, we introduce the entropy and relative entropy for regression problems and classification problems, respectively. The proposed framework is applied to stock market data and an image classification task. As a result, the proposed framework shows superior estimation performance, especially on noisy data; moreover, it is demonstrated that the proposed framework can properly estimate the uncertainty of predictions.
\end{abstract}

\section{Introduction}

Conventional predictive Artificial Neural Network (ANN) models commonly operate with a feed-forward framework using deterministic weight matrices as the network weight parameters \citep{RN1,RN2,RN3}. Specifically, the estimation of ANNs is conducted with matrix operations between given samples and the trained network parameters with non-linear activation functions.

While outstanding progress has been made in ANNs in recent years \citep{RN4,RN5}, however, conventional predictive ANN models have an obvious limitation since their estimation corresponds to a point estimate. Such a limitation causes the restrictions of using ANN for medical diagnosis, law problems, and portfolio management, where the risk of the predictions is also essential in practice. The conventional ANN models produce the same form of predictions even if the predictions are very uncertain; and such uncertain prediction results cannot be distinguished from confident and regular predictions.

In short, conventional ANN models cannot say 'I don't know', and it corresponds to a type of overfitting. For instance, the models attempt to make a confident prediction for an outlier or even complete noise data of which predictions are meaningless and impossible. In such a framework, it is not clear how much the models are sure on their predictions.

To handle such a problem, a probabilistic approach of ANNs, called Bayesian Neural Networks (BNNs), has been introduced \citep{RN6,RN7,RN8}. In BNNs, values of the network weight parameters are not fixed, and instead obtained by a sampling process from certain distributions. Therefore, the prediction of BNN for a given sample differs in each operation. Integrated with the Monte Carlo method in which many different predictions are made for a given sample, the prediction of BNNs also forms a distribution of which variance can represent the risk and uncertainty of the predictions.

However, the training of BNNs is not straightforward since the Monte Carlo method is employed for the training process as well \citep{RN6}. In the training process, the network parameters are sampled from posterior distributions of the network parameters, and gradients are calculated and back-propagated for the sampled parameters. Therefore, such randomness intrinsic in the training process hinders fast training and convergence of BNNs. Furthermore, the training of deep BNNs is also not straightforward, due to such a problem.

As another probabilistic neural network model that can produce a form of distributions as its outputs, Generative Adversarial Networks (GANs) have shown superior performance for sample generation \citep{RN9,RN10,RN11,score,recurrentGAN}. Generally, GANs learn the sample distribution of a certain dataset in order to produce synthetic but realistic samples from input noises, by mapping features intrinsic in the dataset onto the input noise space. While typical GAN models produce random samples, conditional variants of GANs (cGANs) have been introduced to generate desired samples, and have shown fine results to produce samples by using conditional inputs \citep{RN12,RN13,RN14,RN15}.

Basically, this paper addresses the following question that has arisen from the characteristic of cGANs: Since cGANs can learn the probability distribution of samples, i.e., Pr$(X|Y,Z)$, is it possible to learn the probability distribution of labels, i.e., Pr$(Y|X,Z)$, by reversing the inputs and outputs of the cGANs? If it is possible, we can utilize cGANs as a predictive probabilistic neural network model, similar function to BNNs. Also, such a model can solve the problems in BNNs since deep architectures can be employed for GANs, and their training is relatively simple, compared to BNNs. However, such an issue has not yet been studied extensively.

In this paper, we propose an adversarial learning framework for utilizing the generator in cGANs as a predictive deep learning model with uncertainty. Since the outputs of the proposed model are a form of distribution, the uncertainty of predictions can be represented as the variance of the distribution. Furthermore, to measure and quantify the uncertainty of estimations, we introduce the entropy and relative entropy for regression problems and classification problems, respectively.

\section{Background}
\subsection{Problem description}

Let $\bm{X} \in \mathbb{R}^{p}$ be a sample and $\bm{Y}\left(\bm{X}\right) \in \mathbb{R}^{q}$ be labels for the sample. An predictive ANN model estimates the labels for a given sample as follows:
\begin{equation}
\bm{\hat{Y}}_{\mathcal{M}, \bm{\vartheta} _{A}} \left( \bm{X} \right) =\mathcal{M} \left( \bm{X}; \bm{\vartheta} _{A} \right),
\end{equation}
where $\mathcal{M} \colon \mathbb{R}^{p} \mapsto \mathbb{R}^{q}$ is an ordinary ANN structure and $\bm{\vartheta}$ denotes a set of weight parameters. Throughout the paper, we use the same notation $\mathcal{M}_{ \left(\cdot\right) }$, when an ANN structure is used, and $M_{\left(\cdot\right)}$ is used to indicate a general method that does not use ANNs.

However, such predictions correspond to point estimates and it is not clear how much the model is certain of the predictions. For instance, given a complete noise $\bm{X}_{N} \sim N\left(\bm{0},\bm{I}\left(p\right)\right)$, it is even possible to predict a label for the noise, i.e., $\bm{\hat{Y}}_{\mathcal{M}, \bm{\theta} } \left(\bm{X}_{N}\right)$; obviously, such a point estimate is a wrong answer for noises, and one of the correct answer might be ‘I don’t know’, which is impossible to be represented in conventional ANNs.

In this paper, we aim to solve this problem by conducting the prediction in a probabilistic manner:
\begin{equation}\label{eq:2}
\text{Pr} \left( Y_{1} \left( \bm{X} \right), ...,  Y_{q} \left( \bm{X} \right) \right) =\mathcal{M}_{P} \left( \bm{X}; \bm{\vartheta}_{P} \right),
\end{equation}
where $\mathcal{M}_{P} \colon \mathbb{R}^{p} \mapsto \mathbb{D}$ is a probabilistic ANN model, $Y_{j}$ denotes a variable for possible values of $j^{th}$ element of $\bm{Y}\left(\bm{X}\right)$, and $j \in \{1,...,q\}$. From such probability distributions of predictions, not only the point estimation but the uncertainty of the prediction can be calculated as follows:
\begin{equation}
\begin{gathered}
\left\{ \bm{\hat{Y}}_{M_{C},\mathcal{M}_{P}, \bm{\vartheta}_{P}} \left( \bm{X} \right) , \eta  \left( \bm{\hat{Y}}_{M_{C},\mathcal{M}_{P}, \bm{\vartheta}_{P}} \left( \bm{X} \right)  \right)  \right\}
= M_{C} \left( \text{Pr} \left( Y_{1} \left( \bm{X} \right), ...,  Y_{q} \left( \bm{X} \right) \right) ; \bm{\theta} _{C} \right),
\end{gathered}
\end{equation}
where $M_{C} \colon \mathbb{D} \mapsto \mathbb{R}^{2}$ is a method for point estimation using the estimated distributions; therefore, $\bm{\hat{Y}}_{M_{C}}$ can be the mean or median of the estimated distributions, $\bm{\theta} _{C}$ is a set of parameters for $M_{C}$, and $\eta$ denotes a value describing the uncertainty measure of the point estimation.

\subsection{Stochastic weights for neural networks}

To construct a probabilistic neural network model shown in (\ref{eq:2}), BNNs using stochastic weights have been introduced \citep{RN6,RN16,RN17,RN18}. In BNNs, each weight has a probability distribution; and a value of the weight is sampled from the probability distribution each time the model make an inference. Then, by using the Monte Carlo method over the distribution, the prediction of BNNs can be a form of distribution as follows:
\begin{equation}
\bm{\hat{Y}}_{\mathcal{M}_{B}, \bm{\vartheta} _{B}\left(i\right)} \left( \bm{X} \right) =\mathcal{M}_{B} \left( \bm{X}; \bm{\vartheta} _{B}\left(i\right) \right),
\end{equation}
\begin{equation}\label{eq:5}
\begin{gathered}
\text{Pr} \left( Y_{1} \left( \bm{X} \right), ...,  Y_{q} \left( \bm{X} \right) \right)
= M_{D}\left(\left\{\bm{\hat{Y}}_{\mathcal{M}_{B}, \bm{\vartheta} _{B}\left(1\right)} \left( \bm{X} \right),...,\bm{\hat{Y}}_{\mathcal{M}_{B}, \bm{\vartheta} _{B}\left(k\right)} \left( \bm{X} \right)\right\};\bm{\theta} _{D} \right),
\end{gathered}
\end{equation}
where $\mathcal{M}_{B} \colon \mathbb{R}^{p} \mapsto \mathbb{R}^{q}$ is a BNN architecture, $i$ denotes an index for the weight sampling, $k$ denotes the number of sampling, $\bm{\vartheta}_{B} \left(i\right)\sim P_{\bm{\vartheta}_{B}}$ denotes the $i^{th}$ sampled weights from the weight distributions, $M_{D} \colon \mathbb{R}^{k} \mapsto \mathbb{D}$ and $\bm{\theta} _{D}$ are a method for the probability density estimation and a set of parameters for the method, respectively.

\subsection{Generative adversarial networks and their conditional variants}

GANs are probabilistic neural network models in common with BNNs; however, in contrast to BNNs, GANs use deterministic weight parameters, and instead employ a stochastic noise vector as their input for representing latent features in a dataset \citep{RN19,RN20,RN21,RN22,RN23}. The training of GANs is performed by an adversarial manner in which a discriminator and a generator play a game to distinguish and deceive each other.
Generally, GANs are used to learn sample distributions; therefore, varying by the noise vector with the Monte Carlo method, the output of the generator is a synthetic sample:
\begin{equation}
 \bm{\hat{X}}_{\mathcal{M}_{G}, \bm{\vartheta} _{G}} \left( \bm{Z} \left( i \right)  \right) =\mathcal{M}_{G} \left( \bm{Z} \left( i \right) ; \bm{\vartheta} _{G} \right), 
\end{equation}
where $\mathcal{M}_{G}$ is a generator in GANs, $\bm{Z} \left( i \right) \sim P_{Z}$ is a noise vector, and $\bm{\hat{X}}$ is a generated synthetic sample.

However, since it is not clear which noise variable is related to which feature, producing desired samples is challenging in ordinary GANs. To address such a problem, conditional variants of GANs have been studied. Conditional GAN (cGAN), one of the most popular of the conditional variants, is extensively used to generate conditional samples \citep{RN12,RN24,RN25,RN26}. cGAN uses labels as another input for the generator:
\begin{equation}\label{eq:7}
 \bm{\hat{X}}_{\mathcal{M}_{cG}, \bm{\vartheta} _{cG}} \left( \bm{Z} \left( i \right),\bm{Y}  \right) =\mathcal{M}_{cG} \left( \bm{Z} \left( i \right),\bm{Y} ; \bm{\vartheta} _{cG} \right), 
\end{equation}
where $\mathcal{M}_{cG}$ is a generator in cGANs, and $\bm{Y}$ is a desired condition, which is basically the same form as $\bm{Y}\left(\bm{X}\right)$.

\section{Methods}
\subsection{Conditional generative adversarial networks as a prediction model}
In this paper, we propose a new framework to use the generator in cGAN as a predictive model while the existing cGAN is routinely employed for sample generation. By simply reversing the output and the conditional input in cGAN, the model can successfully be used as a probabilistic predictive neural network model which has the same function as BNNs:
\begin{equation}
\begin{gathered}
 \bm{\hat{Y}}_{\mathcal{M}_{cG}, \bm{\vartheta} _{cG}} \left( \bm{Z} \left( i \right) ,\mathcal{M}_{F} \left( \bm{X}; \bm{\vartheta} _{F} \right)  \right)
 =\mathcal{M}_{cG} \left( \bm{Z}  \left( i \right) ,\mathcal{M}_{F} \left( \bm{X}; \bm{\vartheta} _{F} \right) ; \bm{\vartheta} _{cG} \right)
\end{gathered}
\end{equation}
where $\mathcal{M}_{F} \colon \mathbb{R}^{p} \mapsto \mathbb{R}^{u}$is a feature network that extracts $u$-dimensional features from samples, and therefore, $\bm{\hat{Y}}_{\mathcal{M}_{cG}}$ corresponds to one of the prediction results using cGAN; we can use the sample $\bm{X}$, instead of the feature network, if the dimension of the sample space, i.e., $p$, is low. Such a modification is simply changing the input and output in (\ref{eq:7}).

By sampling different noise vectors $\bm{Z} \left( i \right)$, a probability distribution of predictions can be obtained in a similar manner with BNNs as described in (\ref{eq:5}):
\begin{equation}\label{eq:9}
\begin{gathered}
 \text{Pr} \left( Y_{1} \left( \bm{X} \right), ...,  Y_{q} \left( \bm{X} \right) \right) \\
= M_{D}\left(\left\{\bm{\hat{Y}}_{\mathcal{M}_{cG}, \bm{\vartheta} _{cG}} \left( \bm{Z}\left(1\right), \mathcal{M}_{F} \left( \bm{X}; \bm{\vartheta} _{F} \right) \right) ,..., \right. \right.
\left. \left. \bm{\hat{Y}}_{\mathcal{M}_{cG}, \bm{\vartheta} _{cG}} \left( \bm{Z}\left(1\right), \mathcal{M}_{F} \left( \bm{X}; \bm{\vartheta} _{F} \right) \right) \right\};\bm{\theta} _{D} \right).
\end{gathered}
\end{equation}

In the training of GAN structures, the generator is trained by an adversarial manner to deceive the discriminator; therefore, the discriminator is required to be set in order to train $\mathcal{M}_{cG}$. In this paper, the projection discriminator \citep{RN12}, which shows superior performance compared to simple concatenation, is employed for the training of the generator. The architecture of the projection discriminator is as follows:
\begin{equation}\label{eq:10}
\begin{gathered}
\mathcal{M}_{Dis} \left(\bm{X}, \bm{Y} \left( \bm{X} \right); \bm{\vartheta} _{Dis} \right) ={\bf W}_{o} \cdot \mathcal{M}_{ \varphi } \left( \bm{Y} \left( \bm{X} \right) ; \bm{\vartheta} _{ \varphi } \right) 
+\mathcal{M}_{F} \left( \bm{X}; \bm{\vartheta} _{F} \right) ^{T} \cdot \mathcal{M}_{ \varphi } \left( \bm{Y} \left( \bm{X} \right) ; \bm{\vartheta} _{ \varphi } \right),
\end{gathered}
\end{equation}
where $\mathcal{M}_{Dis} \colon \{ \mathbb{R}^{p}, \mathbb{R}^{q} \} \mapsto \mathbb{R}$ denotes the projection discriminator, ${\bf W}_{o} \in \mathbb{R}^{1 \times u}$ is a weight matrix for the output layer of the discriminator, $\bm{\vartheta} _{Dis} = \left\{ {\bf W}_{o}\right\} \cup \bm{\vartheta} _{ \varphi }$, and $\mathcal{M}_{\varphi}:\mathbb{R}^{q}\mapsto\mathbb{R}^{u}$ is an ANN structure with an output dimension of $u$.

To solve a classification problem with the proposed framework, where the input of the discriminator $\bm{Y} \left( \bm{X} \right)$ is a one-hot vector, the $\mathcal{M}_{\varphi}$ can be replaced by a matrix as follows:

\textbf{Proposition 1.} The $\mathcal{M}_{\varphi}$ is equivalent to $\bm{Y} \left( \bm{X} \right) \cdot {\bf W}_{\varphi}$ where ${\bf W}_{\varphi} \in \mathbb{R} ^{q \times u}$, if $\bm{Y} \left( \bm{X} \right)$ is a one-hot vector.

\textit{Proof.} For every one-hot vector $\bm{Y}_{1} \left( \bm{X} \right)=\left[1,0,...,0\right],$ $...,$ $\bm{Y}_{q} \left( \bm{X} \right)=\left[0,0,...,1\right]$, every possible output of $\mathcal{M}_{\varphi}$, i.e., $\mathcal{M}_{ \varphi } \left( \bm{Y}_{1} \left( \bm{X} \right)  ; \bm{\vartheta}_{\varphi}\right),$ $...,$ $\mathcal{M}_{ \varphi } \left( \bm{Y}_{q} \left( \bm{X} \right) ; \bm{\vartheta}_{\varphi} \right)$, is equivalent to $\bm{Y}_{1} \left( \bm{X} \right) \cdot {\bf W}_{\varphi} ,...,\bm{Y}_{q} \left( \bm{X} \right) \cdot {\bf W}_{\varphi}$ since there exist a matrix ${\bf W}_{\varphi}$ that satisfies ${\bf W}_{\varphi} =$ $ \left[ \mathcal{M}_{ \varphi } \left( \bm{Y}_{1} \left( \bm{X} \right) ; \bm{\vartheta}_{\varphi} \right),\right.$ $...,$ $\left. \mathcal{M}_{ \varphi } \left( \bm{Y}_{q} \left( \bm{X} \right) ; \bm{\vartheta}_{\varphi} \right) \right]^{T}$. \QED

Therefore, for classification problems, (\ref{eq:10}) can be simplified as:
\begin{equation}\label{eq:11}
\begin{gathered}
\mathcal{M}_{Dis} \left(\bm{X}, \bm{Y} \left( \bm{X} \right) ; \bm{\vartheta} _{Dis} \right) ={\bf W}_{o} \cdot  \bm{Y} \left( \bm{X} \right) \cdot {\bf W}_{\varphi}
+\mathcal{M}_{F} \left( \bm{X}; \bm{\vartheta} _{F} \right) ^{T} \cdot  \bm{Y} \left( \bm{X} \right)  \cdot {\bf W}_{\varphi}.
\end{gathered}
\end{equation}

Hence, throughout the paper, we use (\ref{eq:10}) and (\ref{eq:11}) for regression problems and classification problems, respectively. 

\begin{figure*}[tp]
\begin{center}
\includegraphics[width=\textwidth]{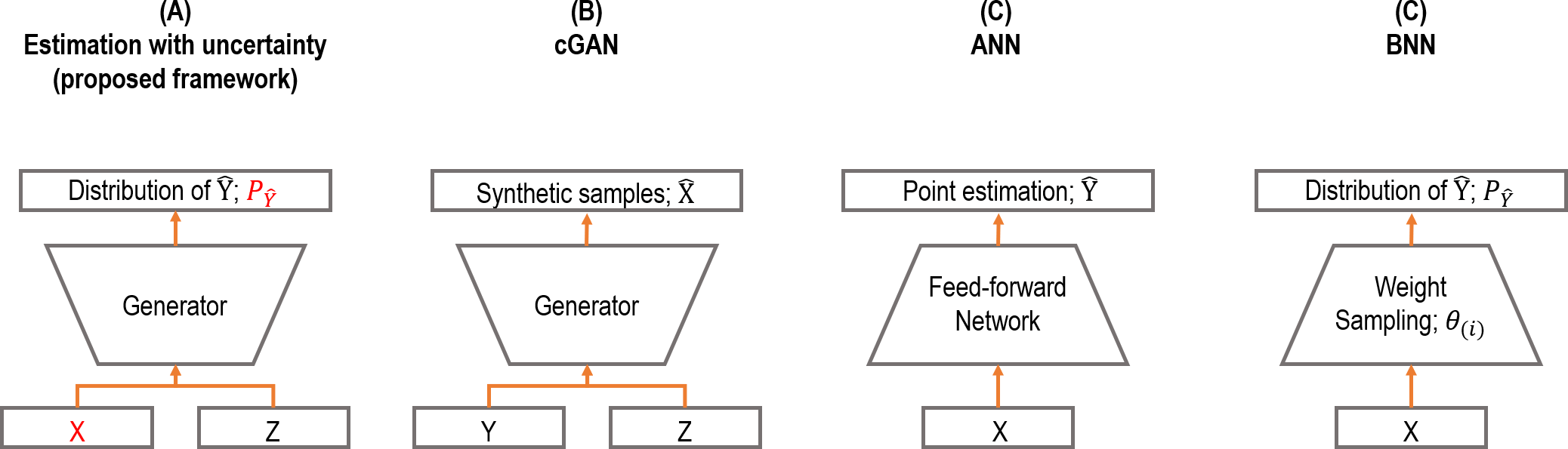}
\caption{\textbf{Comparison to related works. (A)} cGAN as a prediction model (the proposed framework); \textbf{(B)} ordinary cGAN for sample generation; \textbf{(C)} Artificial neural networks for prediction; \textbf{(D)} Bayesian neural networks.}
\label{fig:1}
\end{center}
\end{figure*}

\subsection{Entropy to measure the uncertainty of predictions}
We introduce entropy metrics to measure and quantify the uncertainty of predictions from cGANs. While the variance of estimated distribution can be used to represent the uncertainty if the distribution follows a normal distribution, we employ the entropy in this paper since we cannot reject the possibility that the distribution does not follow a normal distribution. For regression problems, the regular entropy of the estimated distribution in (\ref{eq:9}) is employed, which can be calculated as follows:
\begin{equation}
 H \left( Y_{j} \right) =- \sum _{k} \text{Pr} \left( Y_{j,k} \right)  \cdot \text{log} \left( \text{Pr} \left( Y_{j,k} \right)  \right),
\end{equation}
where $\text{Pr} \left( Y_{j,k} \right)$ the probability of $k^{th}$ element of $Y_{j}$. Notice that $j=1,...,q$, so that the entropy is calculated for each target variable.

On the other hand, for classification problems, the relative entropy, also known as Kullback–Leibler divergence, is used instead of the ordinary entropy since the difference in distributions between the predicted class and the other classes can represent the certainty of the point prediction. If a prediction is certain, the distribution of predicted class has low variance, and its distance from the other distributions would be far; such variance and distance can be comprehensively represented by the relative entropy. Let $l_{X}$ be an index of the predicted class given a sample $\bm{X}$; for example, $l_{X} = \text{argmax}_{j} \sum_{i} \hat{Y}_{j,\mathcal{M}_{cG}, \bm{\vartheta} _{cG}} \left( \bm{Z} \left( i \right) ,\mathcal{M}_{F} \left( \bm{X}; \bm{\vartheta} _{F} \right)  \right)$, if we use the average as the point estimation of each class. Since the relative entropy is asymmetric, we employ a symmetric variant of the relative entropy as follows:
\begin{equation}
\begin{gathered}
 D_{KL} \left( \mathcal{D}_{j=l_{X}}\parallel \mathcal{D}_{j \neq l_{X}} \right) +D_{KL} \left( \mathcal{D}_{j \neq l_{X}}\parallel \mathcal{D}_{j=l_{X}} \right) \\
 = \sum _{k} \text{Pr}_{j=l_{X}} \left( Y_{j,k} \right)  \cdot \mathrm{log} \left( \frac{\text{Pr}_{j=l_{X}} \left( Y_{j,k} \right) }{\text{Pr}_{j \neq l_{X}} \left( Y_{j,k} \right) } \right) \\
 + \sum _{k}^{}\text{Pr}_{j \neq l_{X}} \left( Y_{j,k} \right)  \cdot \mathrm{log} \left( \frac{\text{Pr}_{j \neq l_{X}} \left( Y_{j,k} \right) }{\text{Pr}_{j=l_{X}} \left( Y_{j,k} \right) } \right).
\end{gathered}
\end{equation}

Employing these entropies, we can measure the uncertainty of predictions for regression problems and classification problems as follows:
\begin{equation}
\eta  \left( \bm{\hat{Y}} \right) := \Big\{ \begin{array}{lr}
	 \left\{ H \left( Y_{1} \right) , \ldots ,H \left( Y_{q} \right)  \right\}; & \text{for regression},\\
	\makecell{-D_{KL} \left( \mathcal{D}_{j=l_{X}}\parallel \mathcal{D}_{j \neq l_{X}} \right) \\ -D_{KL} \left( \mathcal{D}_{j \neq l_{X}}\parallel \mathcal{D}_{j=l_{X}} \right)}; & \text{for classification}.\\
	\end{array}
\end{equation}

Notice that $-D_{KL}$ is used to describe the uncertainty since the ordinary relative entropy represents the difference in class distributions; therefore, the minus relative entropy indicates similarity between the predicted class distribution and the other class distributions, which corresponds to the uncertainty. In contrast, the ordinary entropy measures a type of variance of distributions; thereby, a high value of the entropy indicates the uncertainty of predictions.

For classification problems, however, ordinary ANN models have a sort of uncertainty measure of which function is similar to the proposed uncertainty measure. The softmax function is commonly employed for the last layer of ANN classifiers, and the outputs of the function provide a kind of probabilities of each class. Therefore, $\text{log} \left( \bm{\hat{Y}}_{\mathcal{M}, \bm{\vartheta} _{A}} \left( \bm{X} \right)\right)_{l_{X}}$, the cross-entropy loss for the softmax function, can indicate the uncertainty of prediction. However, there exists overfitting in ordinary ANN models as described in the previous section; thereby, this measure becomes imprecise. We will compare the performance between the proposed uncertainty measure and this existing method in Section \ref{sec:res}.

\subsection{Comparison to related works}

In this section, we compare the proposed framework with related prior works. The key differences are summarized and illustrated in Figure \ref{fig:1}.

\textbf{Comparison to ordinary cGANs.} The conventional generator in cGAN is used for synthetic sample generation \citep{RN12}. Ordinary cGAN learns the conditional sample distribution, i.e., $\text{Pr} \left(\bm{X} |\bm{Y},\bm{Z} \right)$, then, produces synthetic samples, integrated with the Monte Carlo method over the noise vector $\bm{Z}$. In contrast, we use cGAN as a prediction model where the model learns the target distribution, i.e., $\text{Pr} \left(\bm{Y} \right)$, and performs predictions as a form of distributions, i.e., $\text{Pr} \left(\bm{Y} |\bm{X},\bm{Z} \right)$, with a stochastic input $\bm{Z}$, given a sample $\bm{X}$ as the conditional input of cGAN. In short, the neural network architecture of cGANs in both studies are basically identical while the proposed framework corresponds to reversing the input and output in the typical use of cGAN.

\textbf{Comparison to ANNs and BNNs.} ANN has a limitation to express the uncertainty of predictions, as described in the previous sections. The output of an ANN is a point estimate, while the output of the proposed framework is a distribution that can represent the uncertainty of predictions, which can handle such a limitation. Likewise, BNN also produces predictions as a form of distribution, which is similar to the proposed framework; however, BNN uses stochastic weights to perform such work, which generally hinders the convergence in training and construction of deep neural network architecture. In contrast, the proposed framework uses deterministic weights and stochastic inputs instead. In addition, the training process is also different, where an adversarial training manner with a discriminator is employed for the proposed framework, which can avoid overfitting resulting from the high complexity of neural network architectures.

\section{Results}\label{sec:res}
\subsection{The prediction of stock prices with the uncertainty measure of the prediction}

Stock market prediction is one of the most specific problems that the prediction of returns and the uncertainty of the prediction are comprehensively required in practice. In the modern portfolio theory \citep{RN27,RN28,RN29,portfolio}, both expected returns and risks of a portfolio must be calculated for the selection of a portfolio; the prediction of the expected returns and the risks exactly corresponds to the estimation with uncertainty, the aim of the proposed framework.

We apply the proposed framework to NASDAQ-100 Future Index data. The model is trained with returns of the past 30 days as the input and the 5-day return as the target. For the training set, the close price index from January 2001 to December 2015 are used; the data from January 2016 to April 2019 are employed for the test set. For the stability of the data, the price data are preprocessed to the returns. The preprocessing process for the price data is provided in Appendix.

The hinge loss and Wasserstein distance are employed for fine training of cGAN, according to recent studies of cGAN \citep{RN12,regular}. For probabilistic models that produce a form of distributions as their output, the mode is employed for the point estimation. The ANN and cGAN-UC, that is cGAN for the prediction with uncertainty, i.e., the proposed framework, are trained over 2,000 epochs; in contrast, BNN models are trained over 5,000 epochs, but the 5-layer BNN fails to converge within the training process, which demonstrates the difficulty of the training of stochastic weights in BNNs, as described previously. The architectures of the models used in this experiment are provided in Appendix.

\begin{table}[tp]
   \caption{\textbf{Correlation coefficients in the test set.} The notation MODEL-n denotes n-layer MODEL. The cGAN-UC indicates the proposed framework, i.e., cGAN for estimation with uncertainty. The means and the standard deviations are obtained over 5 repetitions.} 
   \label{tab:1}
   \centering 
\begin{threeparttable}
   \begin{tabular*}{\linewidth}{l @{\extracolsep{\fill}} ll} 
   \toprule[\heavyrulewidth]\toprule[\heavyrulewidth]
   \textbf{\makecell{Methods}} & \textbf{\makecell{Point estimation \\ vs Target}} & \textbf{\makecell{Uncertainty \\ vs Error}} \\ 
   \midrule
Ridge regression & 0.055 & N/A\tnote{*} \\
Lasso regression & 0.047 & N/A\tnote{*} \\
Random forest & 0.050$\pm$ 0.002 & N/A\tnote{*} \\
ANN-3 & 0.065$\pm$ 0.019 & N/A\tnote{*} \\
ANN-5 & 0.065$\pm$ 0.021 & N/A\tnote{*} \\
ANN-7 & 0.052$\pm$ 0.032 & N/A\tnote{*} \\
BNN-3 & 0.021$\pm$ 0.019 & 0.102$\pm$ 0.081 \\
BNN-5 & N/A\tnote{$\dagger$} & N/A\tnote{$\dagger$} \\
   \midrule
cGAN-UC-3 & 0.076$\pm$ 0.025 & 0.259$\pm$ 0.031 \\
cGAN-UC-5 & \textbf{0.084}$\bm{\pm}$ \textbf{0.052} & 0.278$\pm$ 0.021 \\
cGAN-UC-7 & 0.056$\pm$ 0.076 & \textbf{0.315}$\bm{\pm}$ \textbf{0.061} \\
   \bottomrule[\heavyrulewidth] \bottomrule[\heavyrulewidth] 
   \end{tabular*}
\begin{tablenotes}
\item[*] The uncertainty cannot be calculated with the deterministic models.
\item[$\dagger$] The model failed to converge within 5,000 epochs.
\end{tablenotes}
\end{threeparttable}
\centering 
\end{table}

The models are comprehensively evaluated by the prediction performance of returns and whether the estimated uncertainty is actually correlated with prediction errors, which means the risk of predictions can properly be measured. Table \ref{tab:1} shows correlation coefficients in the test set. The deterministic models show similar prediction performances, whereas cGAN-UCs show superior performance compared to the deterministic models as well as BNNs. The 5-layer cGAN-UC demonstrates the best prediction performance, while the uncertainty estimation is more precise in the 7-layer model. We conjecture that such a performance gain results from the adversarial learning process of GAN that can assist in learning the true distribution of noisy data and avoiding overfitting inherent in the complex neural network architectures, which should be studied further for future work.

\begin{figure*}[tp]
\begin{center}
\includegraphics[width=\textwidth]{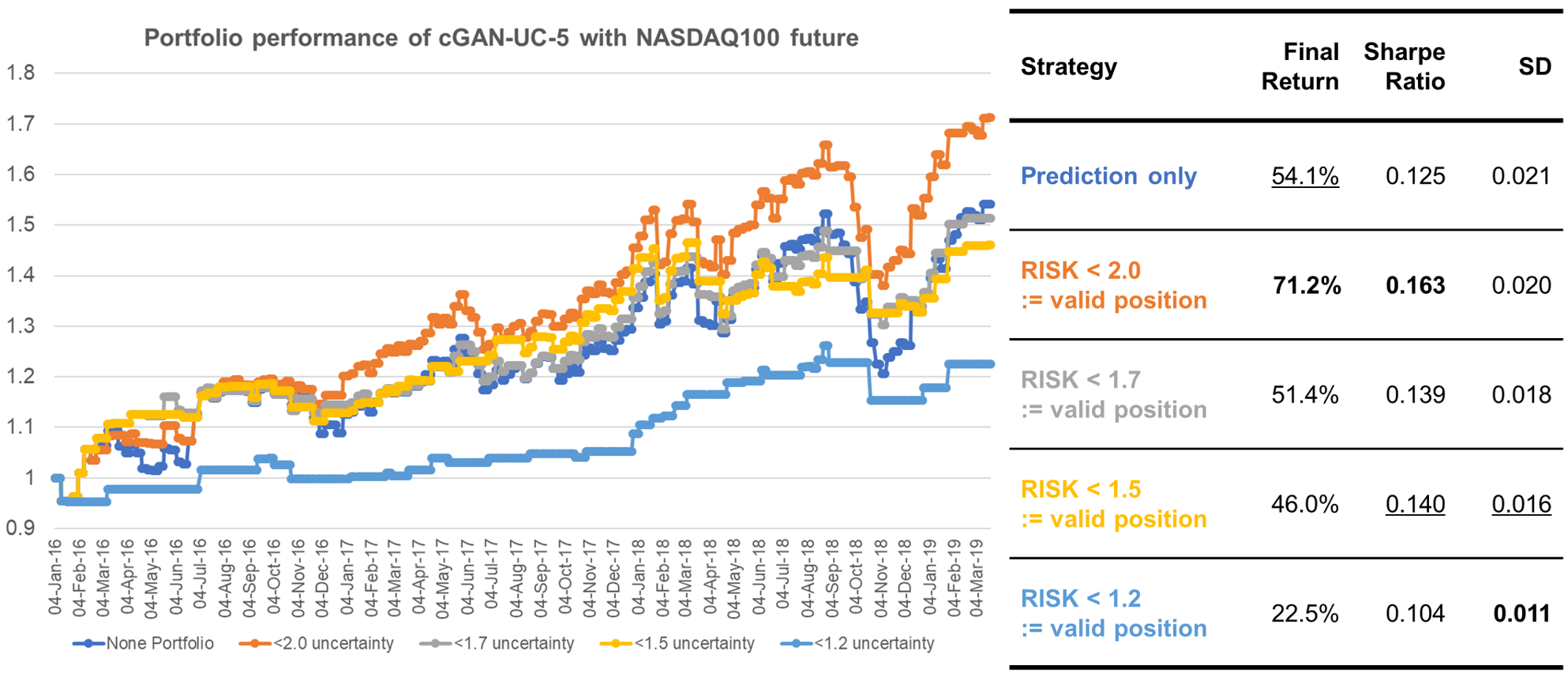}
\caption{\textbf{Portfolio performance with different strategies using cGAN-UC.} Dark blue indicates a conventional strategy using only the prediction of returns while the others represent the strategies using the both, predictions and estimated uncertainty.}
\label{fig:port}
\end{center}
\end{figure*}

Moreover, it is demonstrated that the estimated uncertainty and prediction errors are correlated, which signifies the predictions with high uncertainty have a high possibility of being wrong. While the predictions of returns by both the deterministic models and the probabilistic models are not very successful, in which the correlation coefficients are below 0.1, due to the chaotic nature of the stock market; however, the proposed framework demonstrates that the uncertainty of the predictions can be measured. Such a result indicates a possibility to utilize the proposed uncertainty measure for risk management of portfolios using stock market predictions.

Figure \ref{fig:port} is an example of using the estimated uncertainty for portfolio management. When only the predictions of returns are given, the simplest strategy that uses the predictions is taking a long position if the prediction is positive, and taking a short position otherwise. The portfolio performance of the simplest strategy using only the prediction results of cGAN-UC is represented with dark blue in Figure \ref{fig:port}. We can further enhance the performance, by using the estimated uncertainty (risk), along with the predictions. The other strategies employ the uncertainty by introducing a neutral position if the uncertainty is above certain thresholds; such an approach signifies that we do not take a risk for uncertain predictions. The orange strategy in the figure considers predictions with $>2.0$ uncertainty as invalid predictions; thus, we take a neutral position in such cases. By using the strategy, the final return increases by 17.1\% points, and simultaneously, the standard deviation of the portfolio decreases, compared to the conventional strategy. Furthermore, interestingly, the standard deviation of the strategy with $<1.7$ uncertainty (gray) considerably decreases while the final returns of the strategies are similar to that of conventional strategy. Such a possibility of performance enhancement is another strong evidence that the estimated uncertainty is effective.

\begin{figure}[tp]
\begin{center}
\includegraphics[width=\textwidth]{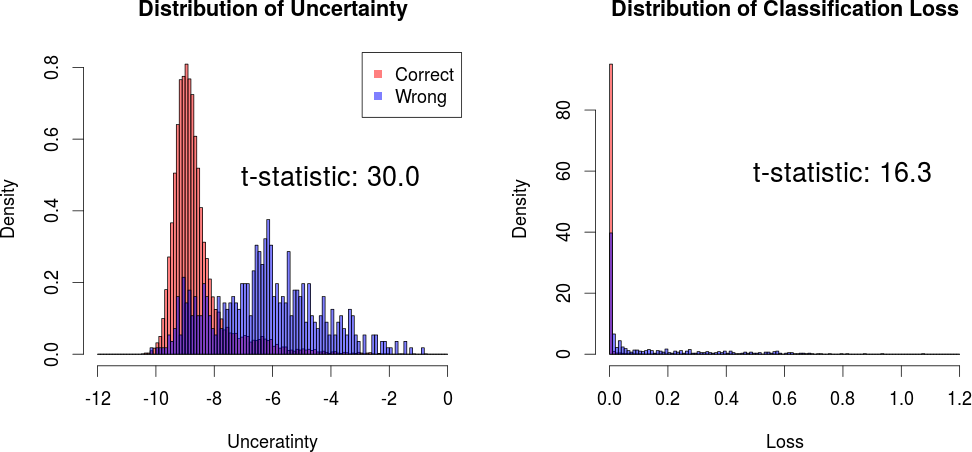}
\caption{\textbf{Comparison between the proposed uncertainty and classification loss in the test set of CIFAR-10.} Light red indicates the uncertainty/loss distributions of which predictions are correct. Blue indicates the uncertainty/loss distribution of which predictions are wrong.}
\label{fig:2}
\end{center}
\end{figure}

\subsection{Image classification with uncertainty}

In this section, the proposed framework is applied to an image classification task with CIFAR-10 dataset. Unlike regression tasks, for classification tasks, deterministic models can estimate the uncertainty of predictions, by using the probability of point estimates as the certainty of predictions, since a softmax function is used for the last layer of the network in general. For instance, if a prediction of a deterministic model is ‘Car’ with a 98\% probability, the uncertainty can be estimated by 2\%. In this manner, for the uncertainty measure of deterministic models, the log of the probability of point estimates, i.e., the cross-entropy classification loss, is employed as a conventional method.

\begin{figure*}[tp]
\begin{center}
\includegraphics[width=\textwidth]{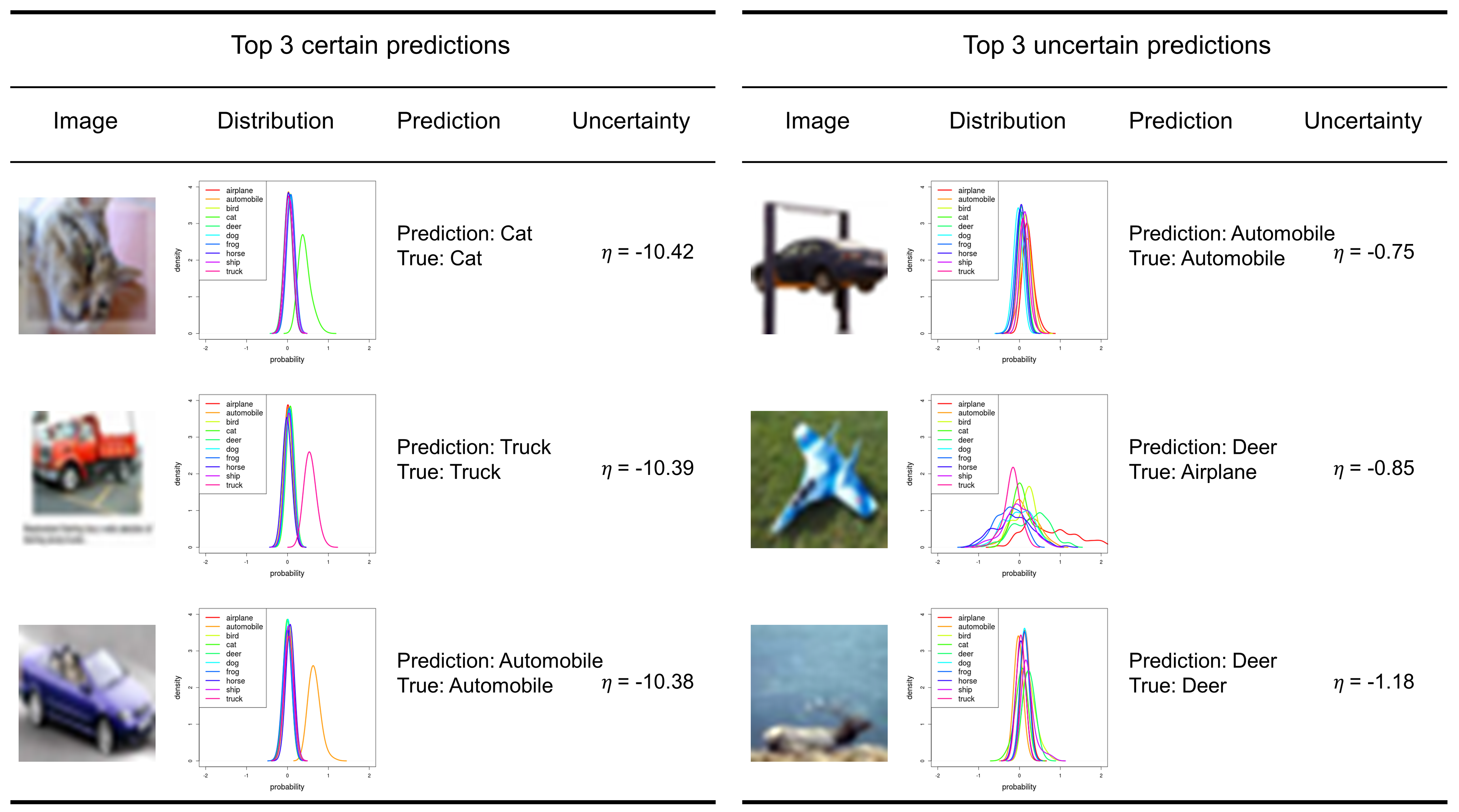}
\caption{\textbf{Certain predictions and uncertain predictions for the test set of CIFAR-10.} Top 3 certain predictions by cGAN-UC (left). Top 3 uncertain predictions by cGAN-UC (right). The 'Distribution' column corresponds to the form of prediction results of cGAN-UC; each color indicates the estimated distribution for each class. Notice that the prediction result in second row of the uncertain predictions is wrong, and the others are correct.}
\label{fig:3}
\end{center}
\end{figure*}

For the comparison between a deterministic model and cGAN-UC, densely connected convolutional networks (DenseNet), which generally shows fine performances for image classification tasks \citep{RN30,RN31}, is used for the feature network of cGAN-UC as well as the deterministic model. The architectures of cGAN-UC and DenseNet used in this experiment are provided in Appendix. As a result, the estimation accuracy of the two models is marginally different, where the test set prediction accuracy of cGAN-UC and ordinary DenseNet is 94.4\% and 94.1\%, respectively, which corresponds to a 0.3\% point of enhancement, by using cGAN-UC.

Moreover, it is demonstrated that there exists a significant performance difference in the uncertainty measures of the models. In short, the conventional method to estimate the uncertainty does not perform well in general. Specifically, there is a sort of overfitting in the deterministic model where the model has a high certainty for wrong answers. For instance, in the test set of CIFAR-10, the median softmax output of DenseNet for the wrong answers is 0.968 (96.8\%), which means that the deterministic classifier is overconfident to the estimations that are actually wrong.

We compare the distributions of the proposed uncertainty measure and the conventional uncertainty measure with respect to correct and wrong predictions. Then, the t-test is performed to measure the difference between the two distribution, which can indicate the correlation between the estimated uncertainty and actual prediction results. As a result, the t-statistics of the proposed uncertainty measure and the classification loss are 30.0 and 16.3, respectively, which signifies superior performance to estimate the uncertainty of predictions (Figure \ref{fig:2}).

We preform a qualitative analysis for the most certain predictions and the most uncertain predictions of cGAN-UC (Figure \ref{fig:3}). We take top three certain/uncertain predictions for the analysis. As a result, while the image samples of the certain predictions are obvious samples, those of the uncertain predictions correspond to a sort of outlier, i.e., an automobile in the air with a door, a toy airplane on the grass, and a deer that is hard to be recognized; also, the airplane is incorrectly classified as 'Deer'. Overall results indicate the proposed framework not only shows superior prediction performance but also can properly measure the uncertainty of the predictions.

\subsection{Noisy image classification with uncertainty}

In this section, we further evaluate the proposed framework with noisy image data since we conjecture that the proposed framework shows robust performance against noise because the adversarial learning process is employed for the training of cGAN-UC. In the adversarial learning process, noisy results that are produced by ordinary samples are rejected by the discriminator; thereby, cGAN-UC learns from the rejections, which makes cGAN-UC robust against noise. The proposed framework is evaluated with noisy CIFAR-10 image data, which are obtained by:
\begin{equation}
\begin{gathered}
\bm{X}_{N_{\left( a \right)}} = \left( 1-a \right) \cdot \bm{X} +  a  \cdot \bm{N},\\
\bm{N} \sim \mathrm{Unif}\left( \left[0,1\right],{\bf I}\left( p \right) \right),
\end{gathered}
\end{equation}
where $a$ indicates a parameter for noises and $\bm{X}_{N_{\left( a \right)}} \in \mathbb{R}^{p}$ denotes a noisy sample. In short, in each experiment, we take different $a$, and then evaluate and compare the classification accuracy of the models. The neural network architecture and the other conditions are the same as those of the previous experiment.

\begin{figure}[tp]
\begin{center}
\includegraphics[width=\textwidth]{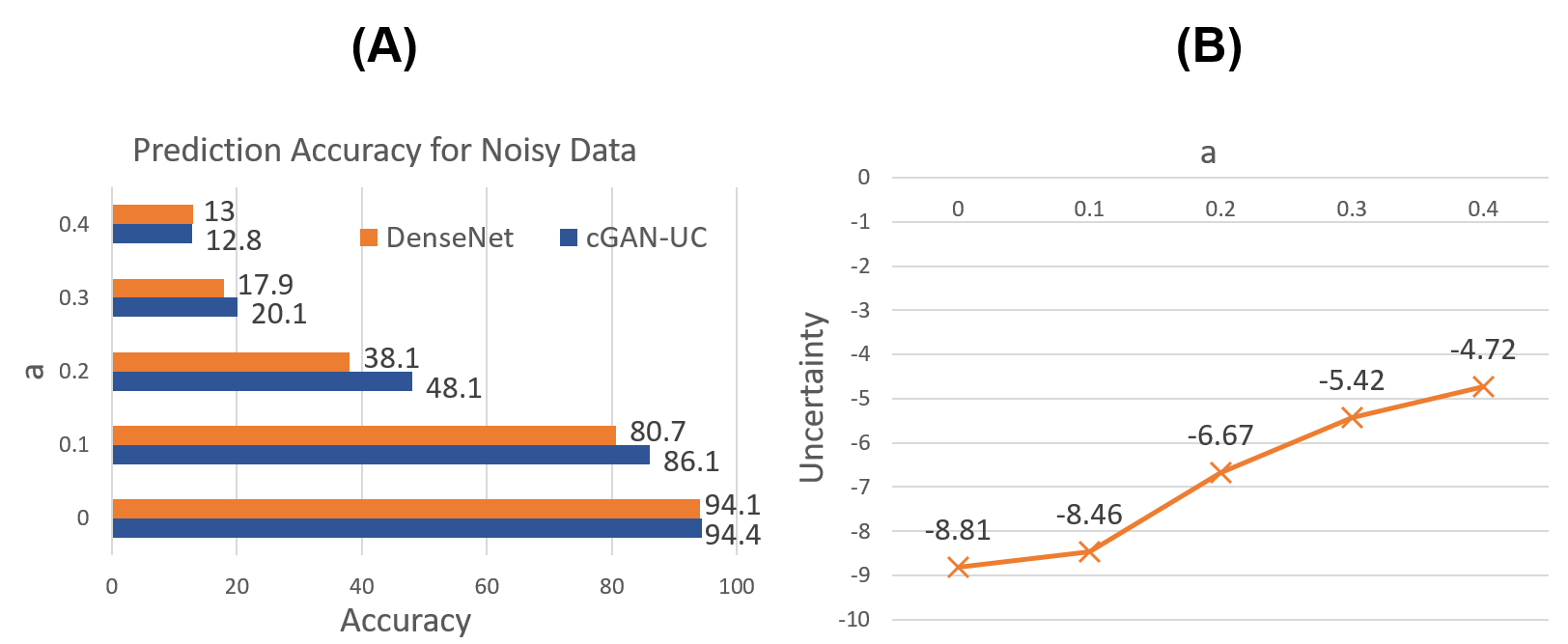}
\caption{\textbf{Prediction results for noisy test set of CIFAR-10. (A)} Prediction accuracy of DenseNet and cGAN-UC w.r.t noise. \textbf{(B)} The proposed uncertainty measure in cGAN-UC w.r.t noise.}
\label{fig:4}
\end{center}
\end{figure}

Figure \ref{fig:4}(A) shows the prediction results for the noisy data. As expected, for not only the original image data but also the noisy data, it is demonstrated that cGAN-UC outperforms the ordinary DenseNet; moreover, the performance difference is more significant in the noisy data. Specifically, with $\bm{X}_{N_{\left( 0.2 \right)}}$, the performance difference is 10.0\% point where the accuracy of cGAN-UC and DenseNet is 48.1\% and 38.1\%, respectively.

In addition, the median of the proposed uncertainty measure is compared with regard to \textit{a}. As shown in Figure \ref{fig:4}(B), the uncertainty increases as the proportion of noises increases. Such a result also strongly supports the claim that the proposed uncertainty measure actually represents the unsureness of predictions.

\section{Conclusion}

We propose a predictive probabilistic neural network framework, which corresponds to a different manner of using the generator in cGAN that is initially introduced for sample generation. While cGAN has commonly been employed for conditional sample generation, with extensive experiments in this paper, it is demonstrated that the model also can be used as a predictive model. In addition, we introduce the uncertainty measures for prediction results of the proposed framework. The uncertainty of prediction is calculated by the entropy and relative entropy for regression problems and classification problems, respectively. The proposed framework was evaluated with stock market data and an image classification task. As a result, the proposed framework demonstrates superior prediction performance and successfully estimates the uncertainty of predictions.

Moreover, interestingly, the proposed framework showed robust performances for noisy data, compared to the deterministic model. We conjecture that these results are due to the adversarial learning process of the proposed framework, where noisy outcomes are rejected by the discriminator, and then the generator learns from the failures. For noisy data, since the performance gain by using the proposed framework is significant, such properties should be investigated further for future work.

In addition, for additional experiments, the proposed framework is evaluated with complete noises, interpolations, and another dataset, and shows superior performance as well. The full results for these additional experiments are provided in Appendix. We expect that the proposed framework can be a significant breakthrough in predictive neural network model, since the proposed framework can predict the uncertainty of estimation that the conventional neural networks can hardly perform, and has a possibility to produce superior performance for prediction tasks. Also, due to the recent developments in GAN training, the proposed framework, compared to BNNs, has advantages in adopting deep architectures and convergence in training process, which have been constant issues in using probabilistic neural networks.

\medskip

\bibliography{uncertain}
\bibliographystyle{IEEEtran}

\end{document}